\documentclass{article}

\PassOptionsToPackage{numbers, compress}{natbib}
\usepackage[preprint]{neurips_2026}

\usepackage[utf8]{inputenc} 
\usepackage[T1]{fontenc}    
\usepackage{hyperref}       
\usepackage{url}            
\usepackage{booktabs}       
\usepackage{amsmath,amssymb,amsfonts}       
\usepackage{nicefrac}       
\usepackage{microtype}      
\usepackage{xcolor}         
\usepackage{graphicx} 
\graphicspath{{figures/}} 
\usepackage{textcomp}
\usepackage{listings}
\usepackage{cleveref}
\usepackage{subcaption}
\usepackage[most]{tcolorbox}
\usepackage{tikz}
\usepackage{wrapfig}

\usepackage{amsthm}
\newtheorem{definition}{Definition}

\newcommand{\col}[1]{%
  \tikz[baseline=(X.base)]{%
    \node[draw=gray!80, fill=gray!10, rounded corners=2pt,
          inner sep=2pt,
          font=\scriptsize\sffamily] (X) {#1};%
  }%
}

\title{Problem Reductions at Scale: Agentic Integration of Computationally Hard Problems}

\author{%
  \textbf{Xi-Wei Pan}\textsuperscript{*\,1} \quad
  \textbf{Shi-Wen An}\textsuperscript{*\,2,\,3} \quad
  \textbf{Jin-Guo Liu}\textsuperscript{1,\,$\dagger$} \\[2pt]
  \textsuperscript{1}\,HKUST (GZ) \quad
  \textsuperscript{2}\,Institute of Science Tokyo \quad
  \textsuperscript{3}\,RIKEN AIP
}

\begin{document}

\maketitle
\begingroup
  \renewcommand\thefootnote{}%
  \footnotetext{\textsuperscript{*}Equal contribution.\quad \textsuperscript{$\dagger$}Corresponding author: \texttt{jinguoliu@hkust-gz.edu.cn}.}%
\endgroup

\begin{abstract}
  Solving an NP-hard optimization problem often requires reformulating it for a specific solver---quantum hardware, a commercial optimizer, or a domain heuristic.
  A tool for polynomial-time reductions between hard problems would let practitioners route any supported problem to any supported solver through a single interface.
  Building such a library at scale, however, has remained out of reach.
  We show that \emph{harness engineering}, the practice of designing constraints, verification systems, and feedback loops that channel AI coding agents, can overcome this barrier.
  Our harness combines a no-code contribution route for domain experts, a multilayer verification stack ranging from type-level checks to agentic feature tests (AI agents role-playing as end users), and a fully automated implementation-review-integration pipeline.
  In about three months, we built a command-line tool backed by a library of 100+ problem types and 200+ reduction rules in over 170k lines of Rust.
  The result suggests that a well-engineered harness lets agents build well-tested software at a scale and pace beyond prior reduction-library efforts.
  Because the reduction graph composes transitively, a new solver registered for any single problem type instantly becomes available to every problem connected by a reduction path.
  The source code is available at \url{https://github.com/CodingThrust/problem-reductions}.
\end{abstract}

\section{Introduction}\label{sec:intro}

\subsection{Many Hard Problems, Few Solvers}

Combinatorial optimization problems arise throughout science and engineering~\cite{korte2008combinatorial,Schuetz2022PhysicsGNN}.
An airline needs to assign crews to flights~\cite{gopalakrishnan2005airline}.
A chip designer needs to allocate registers, commonly modeled as graph coloring in compilers~\cite{chaitin1981register}.
A logistics company needs to route delivery trucks~\cite{toth2014vehicle}.
Each of these is an instance of an NP-hard problem~\cite{karp1972,garey1979}, a class of problems for which no efficient general-purpose algorithm is known, but which can be solved in practice for moderate sizes by specialized solvers.

The difficulty is that each solver speaks its own narrow language.
SAT solvers~\cite{Een2003MiniSat, Biere2020Kissat} accept Boolean satisfiability instances.
Integer linear programming (ILP) engines like CPLEX~\cite{IBM2022CPLEX}, Gurobi~\cite{Gurobi2026}, and HiGHS~\cite{Huangfu2018HiGHS} require problems expressed as linear constraints over integer variables.
Semidefinite programming (SDP) relaxations tackle Max-Cut and graph bisection~\cite{GoemansWilliamson1995} but need a matrix formulation.
Beyond classical solvers, specialized hardware introduces additional formulations: D-Wave quantum annealers~\cite{glover2019} solve Quadratic Unconstrained Binary Optimization (QUBO), and Rydberg atom arrays~\cite{lucas2014, pichler2018, ebadi2022quantum} natively solve Maximum Independent Set on geometric graphs.
Physics-inspired methods scale further: graph neural networks solve QUBO at million-variable scale~\cite{Schuetz2022PhysicsGNN}, and hybrid quantum-GNN methods tackle the Traveling Salesman Problem~\cite{He2024QuantumTSP}.
These approaches still require a QUBO or Ising formulation as input.
A practitioner with a crew-scheduling problem cannot reach most of these solvers without first \emph{translating} the problem into a form the solver understands.

This translation is called a \emph{reduction}: an efficient algorithm that converts an instance of one problem into an instance of another and maps the
target's solution back to the source.
In other words, a reduction gives a uniform rule for solving any source instance via the corresponding target instance.
Each verified reduction is a bridge connecting a new problem to an existing solver.

\subsection{Bridging Problems to Solvers}

\begin{wrapfigure}{r}{0.4\textwidth}
  \vspace{-\baselineskip}
  \centering
  \includegraphics[width=0.4\textwidth]{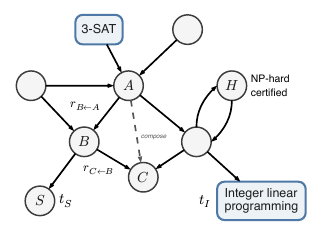}
  \caption{The reduction graph.  Two anchor problems are highlighted: \emph{3-SAT} at the top and \emph{Integer linear programming (ILP)} at the bottom right.  Edge labels are reduction overheads $r_{\cdot \leftarrow \cdot}$; node labels like $t_S$, $t_I$ denote the time complexity of the node's best known solver.}
  \label{fig:reduction-graph-concept}
  \vspace{-\baselineskip}
\end{wrapfigure}

A single reduction between two problems is a \emph{primitive reduction rule}.
Primitive rules collectively form a directed \emph{reduction graph} (\Cref{fig:reduction-graph-concept}): each node is a problem type, each directed edge is a primitive rule.
Every edge carries a \emph{reduction overhead}: a multivariate polynomial mapping the source's size measures to the target's; we write $r_{B \leftarrow A}$ for the overhead of the rule $A \to B$.

Reductions \emph{compose} along directed paths.
Chaining $A \to B$ and $B \to C$ along the central chain in \Cref{fig:reduction-graph-concept} yields a composite rule $A \to C$ with overhead $r_{C \leftarrow A} = r_{C \leftarrow B} \circ r_{B \leftarrow A}$.
Two kinds of paths play a special role, marked by the highlighted nodes in the figure.
3-SAT serves as the source of NP-hardness proofs: any path from 3-SAT to a problem~$H$ provides an NP-hardness argument for~$H$.
We say $H$ is \emph{NP-hard certified} by the reduction graph.
Integer linear programming (ILP) is backed by mature solvers: any path from a problem to ILP routes it to one of them, which is crucial for testing the correctness of reductions.
The target solver's time complexity, $t_I$ for ILP, thereby upper-bounds the end-to-end solving cost of any problem routed to it, up to reduction overhead.

\subsection{The Challenge}
\begin{wrapfigure}{r}{0.4\textwidth}
  \vspace{-\baselineskip}
  \centering
  \includegraphics[width=0.3\textwidth]{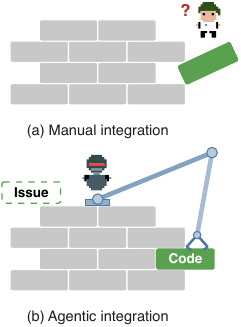}
  \caption{The integration problem.
    (a) A new reduction does not fit the library's conventions and cannot be integrated without learning the codebase.
    (b) The contributor files a structured issue; an agent handles implementation, review, and documentation.}
  \label{fig:comic}
  \vspace{-\baselineskip}
\end{wrapfigure}

Scaling an \emph{executable} reduction graph is an engineering problem.
The literature is far ahead of any implementation: Garey and Johnson~\cite{garey1979} catalogue over 300~NP-hard problems with reduction sketches, yet the largest executable reduction graph (Appendix~\ref{sec:related}) covers fewer than 20 problem types.
Implementation and maintenance are the bottleneck, and two barriers in particular emerge.

\textbf{Convention drift.}
Every reduction rule must conform to the same interface, the same file layout, and the same test pattern.
Maintaining this uniformity across many contributions is precisely what human teams cannot do at scale: as contributors accumulate their own patterns, the codebase fractures.
A mathematician who understands a new reduction has no way to contribute it without first learning the implementation language and the project's conventions (\Cref{fig:comic}).

\textbf{Effort exhaustion.}
Every new entry demands the same lengthy checklist.
Adding a problem type requires resolving whether the problem is genuinely new or a known problem under a different name, writing a formal definition, and identifying the best known algorithm and its complexity.
Adding a reduction rule requires confirming that both endpoint problems exist, verifying the rule is novel and actually improves on existing routes, analyzing the reduction overhead, providing a correctness argument, implementing the forward and inverse maps, designing correctness tests, and documenting the reduction with a worked example.
Repeating this cycle for every entry exhausts even dedicated maintainers.

AI coding agents offer a way forward, but only when paired with a \emph{harness}: the engineered constraints, feedback loops, and automatic checks that channel agents toward reliable output~\cite{OpenAI2026HarnessEngineering}.
The engineer's job is no longer to write the code, but to design the harness that makes the agent's code trustworthy.
We build a domain-specific harness that addresses both \emph{convention drift} and \emph{effort exhaustion}; \Cref{sec:method} details its components.

\subsection{Contributions}

\begin{figure}
  \centering
  \includegraphics[width=\columnwidth]{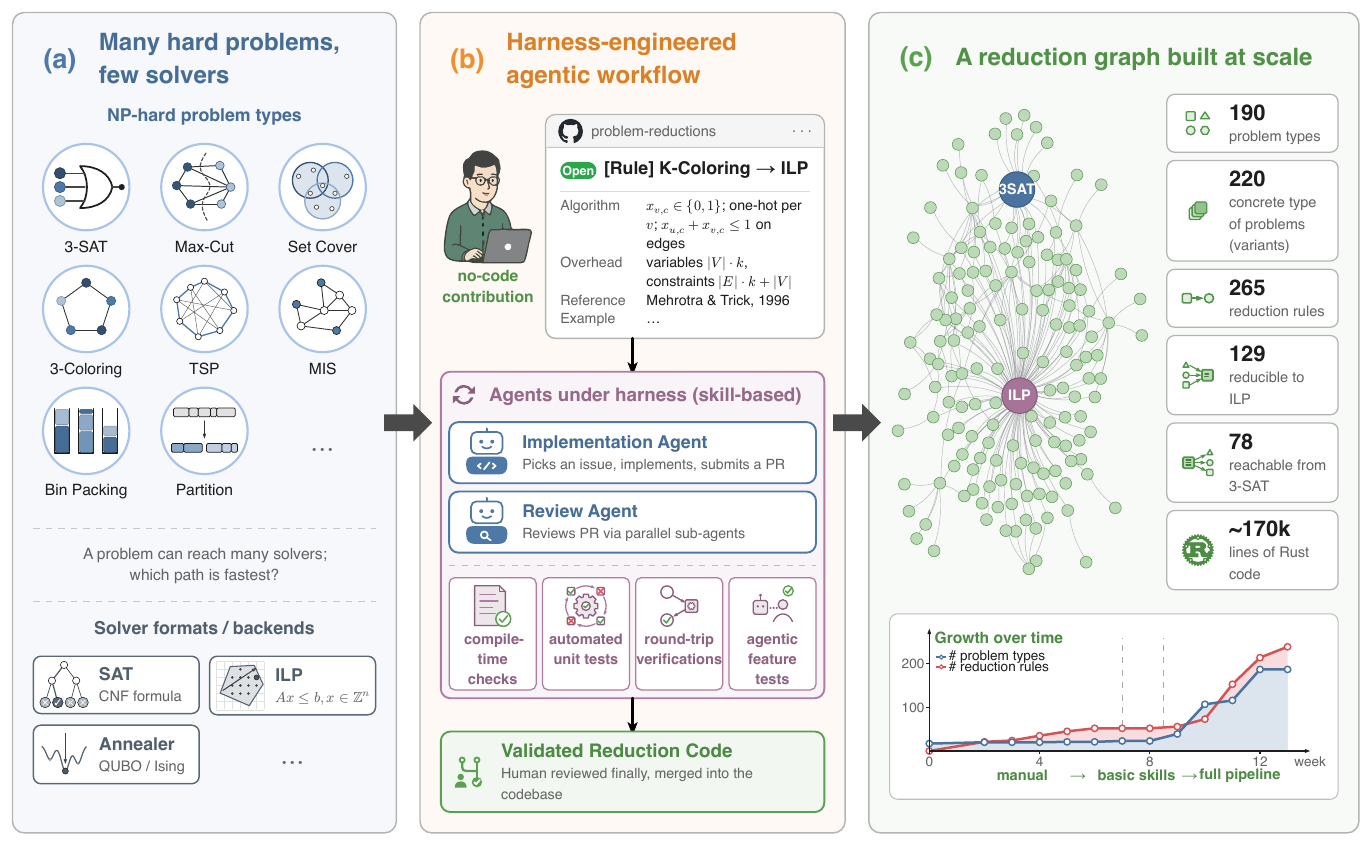}
  \caption{Overview.
    (a) Many solvers exist, and a problem can usually reach several of them through different reductions, but this knowledge is scattered across the literature without a unified executable framework.
    (b) A harness-engineered agentic workflow turns a structured GitHub issue into validated code through implementation and review agents under a layered verification stack (the four main layers shown; \Cref{sec:verification} lists all six).
    (c) Three months of this pipeline produced a reduction graph with 190 problem types and 265 rules; the inset shows growth over time.}
  \label{fig:overview}
\end{figure}

We make three contributions, each addressing one actor in the workflow. 

\textbf{For end users.}
The first comprehensive and executable library of computationally hard problem reductions: 265 reductions among 190 problem types and 170k lines of Rust code, an inventory that would have taken years by hand but landed in three months with a small team and AI agents.

\textbf{For maintainers.}
Our harness uses skills to guide agents through the full pipeline, from coding to the final review, under a layered correctness gate.
Maintainers are freed from routine work to focus on the intellectual work.

\textbf{For domain experts.}
A no-code contribution route that turns a structured GitHub issue into validated implementation, with no familiarity with the implementation language needed.

\begin{figure}[t]
  \centering
  \includegraphics[width=0.85\columnwidth]{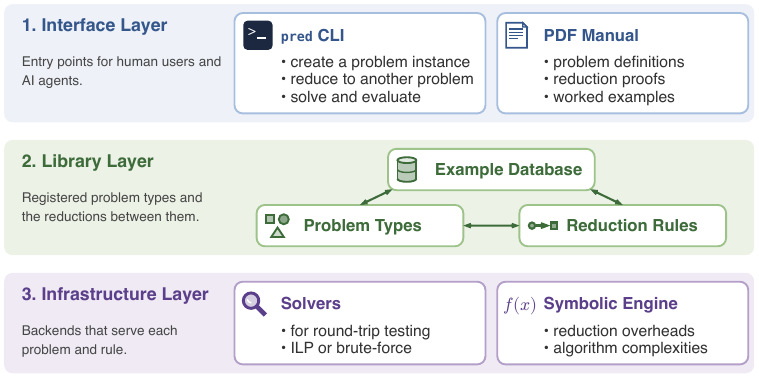}
  \caption{System architecture.
    Three layers separate concerns: the interface layer (top) is what users and agents touch, the library layer (middle) holds the reduction graph itself, and the infrastructure layer (bottom) provides the computational backends: solvers for instances and a symbolic engine for overhead and complexity.}
  \label{fig:library-architecture}
\end{figure}

\section{System Architecture}\label{sec:architecture}

We implement the reduction graph as a Rust crate, organized in three layers (\Cref{fig:library-architecture}).

\subsection{Interface Layer}\label{sec:user-interfaces}

The library exposes two interfaces, both designed for humans and AI agents: the \texttt{pred} command-line interface (CLI) and a PDF manual.

Through the CLI, \texttt{create} emits a problem instance as JSON, \texttt{reduce} transforms it along any reduction path, and the resulting JSON pipes into external solvers or downstream tools.
\texttt{solve} closes the loop for quick checks.
See Appendix~\ref{app:cli} for a worked session.

Compiled from Typst, the manual collects the library's formal mathematics: each problem and reduction has a definition, a worked example, a diagram, and a proof sketch.
Domain experts review it for correctness, and agents register a new entry whenever they add a model or rule, keeping code and mathematics in lockstep.

\subsection{Library Layer}\label{sec:graph-structure}

The library layer is a collection of problem types and the reduction rules between them.

\textbf{Problem types.}
Every problem type implements a uniform interface for problems, such as input domain, solution type and \emph{size measures}.
Each type defines its own size measures and \emph{best-known algorithm complexity}, for example, Maximum Independent Set (MIS) is parameterized by $|V|$ and $|E|$, with a known $O(1.1996^{|V|})$ algorithm~\cite{Xiao2017MIS}.
A problem type may have \emph{variants}: restricting the input domain to a narrower class preserves the combinatorial objective but can change the complexity.
Each variant occupies a distinct node in the reduction graph.
For instance, general MIS requires $O(1.1996^{|V|})$ time, whereas MIS on unit disk graphs admits a $2^{O(\sqrt{|V|})}$ algorithm, which is subexponential, unlike the general case~\cite{deBerg2020ETHTight}.

\textbf{Reduction rules.}
Each rule requires two mappings, a forward map and an inverse map. The forward map converts a source problem instance to a target problem instance.
After solving the target problem, the inverse map extracts source solutions from target solutions.
Each rule also carries a \emph{reduction overhead}: a set of multivariate polynomials mapping the source's size measures to the target's.
For example, the 3-SAT $\to$ MIS reduction creates one vertex per literal occurrence in the formula, yielding $O(L)$ vertices and $O(L^2)$ edges where $L$ is the total number of literal occurrences.

\textbf{Example database.}
Each problem type defines a canonical instance, such as the Petersen graph for Maximum Independent Set.
The same instance drives the test suite and serves as the worked example in the PDF manual.

\subsection{Infrastructure Layer}\label{sec:infrastructure}

The infrastructure layer provides the computational backends that the library layer builds on: backend solvers for exact verification, and a symbolic engine for analyzing computational complexity. Appendix~\ref{app:architecture} presents the engineering details of the Rust crate.

\textbf{Solvers.}
Every problem type needs a minimal but reasonably efficient solver, because testing requires solving each problem exactly.
Most types (e.g., Maximum Independent Set, Traveling Salesman, flow-shop scheduling) have a reduction path to ILP.
The default validation strategy reduces the problem to ILP and solves it with HiGHS~\cite{Huangfu2018HiGHS}, an open-source solver.
For problems without an ILP path, the library provides customized exact solvers that exploit problem structure. For example, functional-dependency closure for Minimum Cardinality Key~\cite{Lucchesi1978CandidateKeys}, and cycle enumeration with branch-and-bound for Partial Feedback Edge Set~\cite{Baharev2021FeedbackArcSet}.
The remaining problems fall back to brute-force enumeration on small instances.

\textbf{Symbolic engine.}
Both reduction overheads and algorithm complexities are represented as symbolic expressions.
The symbolic engine supports three operations: \emph{composition} substitutes one edge's output polynomial into the next edge's input, often used to determine the composite overhead of a multi-step reduction path; \emph{comparison} determines whether a symbolic expression has lower asymptotic growth than another, used for finding the cheapest reduction path; and \emph{evaluation} evaluates a symbolic expression.

\subsection{Why Rust?}\label{sec:why-rust}

None of the maintainers had written Rust before this project.
The language was chosen in large part for properties that benefit agents.

\textbf{Strict compiler, actionable errors.}
Rust rejects many invalid programs before execution, and reports precise type, ownership, and lifetime errors.
These diagnostics give agents a concrete repair target.

\textbf{Fast feedback.}
Cargo provides a standard loop for building, testing, and formatting.
Incremental compilation and unit tests make that loop fast enough for agent iteration.

\section{Harness Design}\label{sec:method}

\subsection{Anatomy of the Harness}\label{sec:skills}

\begin{definition}[Harness System]
A \emph{harness system} constrains agent behavior through five components: a project specification auto-loaded into every agent session, a set of tools the agent may invoke, a progressively-disclosed knowledge base, a set of automation skills (executable checklists that compose the previous three into multi-step pipelines), and a set of advisor skills for human contributors onboarding.
\end{definition}

The five components address successive needs in an agent session: what conventions to follow, what actions to take, what domain knowledge to consult, and how to orchestrate these into workflows.

\textbf{Project specification} files at the repository root (\texttt{CLAUDE.md}, \texttt{AGENTS.md}) encode how to find the remaining four components and what conventions to follow.
Every agent session loads these files automatically on startup, ensuring that a new agent inherits the same conventions as every prior session, without human briefing.

\textbf{Tools} include command-line interfaces (CLI), model context providers (MCP), and external search services.
The CLI and the language's built-in test harness are the agent's hands.
In this project, the \texttt{pred} CLI lets agents create problem instances, run reductions, invoke solvers, and inspect topological connectivity in the reduction graph.
ArXiv MCP and web search retrieve references, and Typst compiles the PDF manual and its documentation artifacts.

\textbf{Knowledge base.}
Domain knowledge includes Garey and Johnson's catalogue~\cite{garey1979}, reference papers converted to Markdown for agent consumption, and reference implementations already in the codebase.
This body of material is too large to fit in a single agent context window.
The harness organizes this knowledge for \emph{progressive disclosure}~\cite{Nielsen2006ProgressiveDisclosure}: agents discover relevant files through search and read full content (formal definitions, proof sketches, worked examples) only when a specific task demands it.

\begin{figure}[t]
  \centering
  \includegraphics[width=\columnwidth]{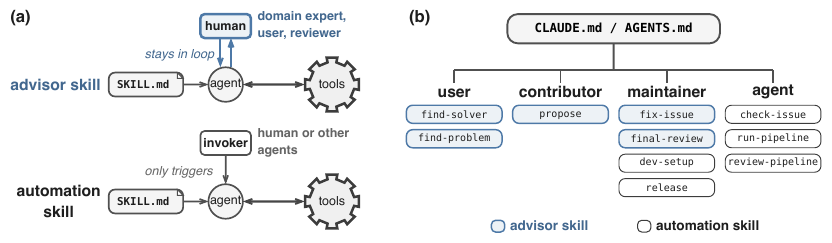}
  \caption{Skills architecture.
    (a) The two skill kinds: \emph{advisor} skills keep humans in the loop with the agent throughout, while \emph{automation} skills run end-to-end once invoked. Other interactions with agents are omitted for clarity.
    (b) The skill catalogue, grouped by invoker (user, contributor, maintainer, or agent). Blue boxes are advisor skills; white boxes are automation skills.}
  \label{fig:skill-architecture}
  \end{figure}

Skills are plain-Markdown documents that orchestrate specification, tools, and knowledge into multi-step, repeatable workflows.
We divide skills into two kinds (\Cref{fig:skill-architecture}): \emph{automation skills} that agents execute end-to-end, and \emph{advisor skills} that bring humans into the loop at decision points.

\textbf{Automation skills} can include \emph{abstract} steps that traditional automation (Makefiles, shell scripts) cannot express: like ``implement the reduction algorithm''(\texttt{run-pipeline} skill in~\Cref{fig:skill-architecture}(b)), ``review the pull request for correctness''(\texttt{review-pipeline} skill) and ``make a new release''(\texttt{release} skill).
These steps require judgment and are interpreted by the agent using its own capabilities.
Unlike bare prompts, skills break a complex task into concrete, ordered steps small enough for an agent to execute faithfully, preventing the drift and omissions that occur when agents plan from scratch.

\textbf{Advisor skills} involve humans for two purposes: \emph{training} humans to use the library or contribute to it, and \emph{utilizing} humans for tasks agents cannot reliably perform, such as critical decisions, hard mathematical reasoning, and creative work like designing canonical examples.
\emph{Training} skills flatten the learning curve:
end users invoke \texttt{find-solver} skill (\Cref{fig:skill-architecture}(b)) to go from a real-world problem to a concrete solving strategy by exploring the reduction graph, or \texttt{find-problem} to discover what problems a given solver covers; contributors invoke \texttt{propose} skill to brainstorm a new problem or reduction in mathematical language and file a well-formed GitHub issue, with no programming required.
\emph{Utilizing} skills retain humans for critical correctness work: maintainers invoke \texttt{fix-issue} to identify and address quality defects such as unsound proofs or low-quality examples interactively, and \texttt{final-review} to catch dangerous commits or correctness gaps before merge.
These skills lower the usage barrier from ``knows the library'' to ``knows the problem,'' and the contribution barrier from ``knows the programming language'' to ``knows the mathematics.''

Appendix~\ref{app:skills} excerpts one skill of each kind: \texttt{propose} (advisor) and \texttt{add-rule} (automation).

\begin{wrapfigure}{r}{0.45\textwidth}
  \vspace{-\intextsep}
  \centering
  \includegraphics[width=\linewidth]{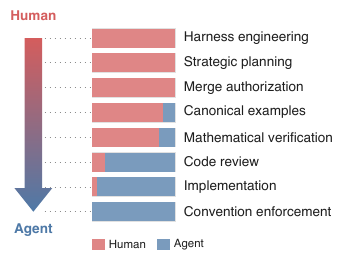}
  \caption{Delegation spectrum. Tasks sorted from fully human (top) to fully automated (bottom); bars show the human--agent split.}
  \label{fig:delegation-spectrum}
\end{wrapfigure}

\subsection{What Cannot Be Delegated}\label{sec:human-responsibilities}

Four responsibilities must remain with humans: \emph{selecting} which reduction to pursue, \emph{verifying} the algorithm's correctness, \emph{constructing} canonical examples, and \emph{authorizing} merges.
Selecting which rules to pursue requires understanding the needs of the community; the human is the information source.
Verifying the algorithm's correctness requires deep reasoning.
Reasoning ability of large language models degrades sharply with problem complexity~\cite{Mirzadeh2025GSMSymbolic,Shojaee2025IllusionOfThinking,Dziri2023FaithFate}. Research-level mathematics remains largely unsolved~\cite{Glazer2024FrontierMath,Paster2026HLE,Merrill2024ExpressivePower}.
Constructing canonical examples requires creative work; the human is the end user and the best judge.
Authorizing merges is security critical; a human with long-term memory must be responsible.

\Cref{fig:delegation-spectrum} summarizes the division of labor, sorted by how replaceable each task is by agents.

\subsection{Automation Pipeline}\label{sec:pipeline-overview}

\begin{figure*}[htbp]
  \centering
  \includegraphics[width=\textwidth]{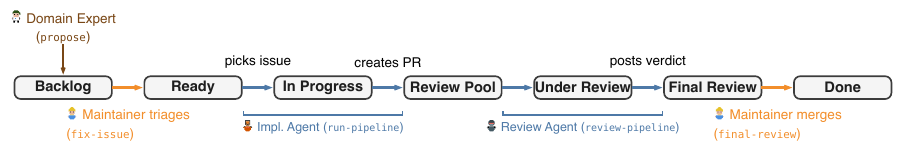}
  \caption{Contribution pipeline as a state machine on the GitHub project board.
    Each box is a board column (state); arrows show the single allowed transition direction.
    Four roles drive the flow: a \emph{domain expert} files issues (brown), a \emph{maintainer} triages and merges (orange), an \emph{implementation agent} picks issues and creates PRs (blue), and a \emph{review agent} runs parallel reviews and posts verdicts (teal).
    If any stage fails, the item moves to On~Hold (not shown) for human triage.}
  \label{fig:pipeline}
\end{figure*}

Adding a reduction to the graph involves six stages; the first five are mirrored as columns on a GitHub project board (\Cref{fig:pipeline}), while Stage~6 runs as a post-merge community call.
Each column is a state; progress flows left to right.
If any stage fails, the item moves to \col{OnHold} for human triage.
Four roles drive the flow: a domain expert files issues, a maintainer triages and merges, and two agents (one for implementation, one for review) handle the routine work in between.

\textbf{Stage~1: Propose} ( $\to$ \col{Backlog}).
A domain expert clones the repository and invokes the \texttt{propose} skill (Appendix~\ref{app:skills}) in Claude Code or a similar agentic coding tool.
The skill conducts a structured brainstorming session (one question at a time) to elicit the core content of the contribution: which problem or reduction rule to add, its formal definition, etc. The expert answers in mathematical language; no programming is required.
The agent runs a topology analysis to avoid duplicates and to identify high-value gaps in the reduction graph, then pre-validates the draft against Stage~2's quality checks.
The output is a structured GitHub issue filed into \col{Backlog}.

\textbf{Stage~2: Validate} (\col{Backlog} $\to$ \col{Ready}).
A maintainer invokes the \texttt{fix-issue} skill to triage each issue in \col{Backlog}.
The agent first evaluates the proposal along four dimensions: \emph{usefulness} (does a cheaper path already exist?), \emph{effort} (what is the effort to implement this rule?), \emph{correctness} (do cited references support the claims?), and \emph{writing quality} (are all symbols defined, all examples fully worked?).
It then auto-corrects mechanical failures (undefined symbols, missing metrics) and brainstorms substantive fixes with the maintainer interactively.
Only issues passing all four checks move to \col{Ready}.

\textbf{Stage~3: Implement} (\col{Ready} $\to$ \col{Review Pool}).
An implementation agent reads the \texttt{run-pipeline} skill, picks an issue automatically from \col{Ready} column, implements it\footnote{The engineering workflow builds on Superpowers~\cite{Obra2025Superpowers}}, submits a pull request (PR) and moves the issue to \col{Review Pool}. All of this happens in headless mode, with no human in the loop.

\textbf{Stage~4: Review} (\col{Review Pool} $\to$ \col{Final Review}).
Similarly, a review agent reads the \texttt{review-pipeline} skill, claims an issue from \col{Review Pool}, moves it to \col{Under Review}, dispatches three read-only sub-agents: structural completeness, code quality, and agentic feature testing (AI agents role-playing as end users; \Cref{sec:verification}), and posts a review report on the PR and moves the issue to \col{Final Review}. All in headless mode.

\textbf{Stage~5: Merge} (\col{Final Review} $\to$ \col{Done}).
A maintainer invokes the \texttt{final-review} skill to review the PR.
Agent claims a PR from \col{Final Review}, checks the review report, performs mechanical fixes, walks the maintainer through the agentic verdict, and lets the maintainer make the final decision.
The maintainer either merges the PR to main branch (issue moved to \col{Done}) or holds the PR with a reason (issue moved to \col{OnHold}).

\textbf{Stage~6: Verify.}
The maintainer invites the domain expert who filed the original issue to a community call, and
walks through the PDF manual and CLI.
This closing-the-loop step catches errors invisible to automated tests: subtle misinterpretations of the problem statement or proof arguments.

Stages~3--4 (implement and review) run in \emph{headless mode} via OpenAI Codex~\cite{OpenAI2026Codex} (\texttt{gpt-5.4}, xhigh reasoning effort).
The remaining stages use Claude Code~\cite{Anthropic2025ClaudeCode} (\texttt{opus-4.6}) for interactive sessions with the maintainer and domain expert.
Anecdotally, we found \texttt{gpt-5.4} more reliable for code implementation and \texttt{opus-4.6} stronger in interactive dialogue, though we did not evaluate this systematically.

\subsection{Correctness Assurance}\label{sec:verification}

Correctness is enforced by six layers: issue review (Stage~2), compile-time type checks, unit tests, round-trip tests, \emph{agentic feature tests} (Stage~4) and manual verification (Stage 5 and 6).

\textbf{Unit tests and round-trip tests.}
Unit tests cover individual problems, solvers, and helper functions.
Round-trip tests cover each reduction rule end-to-end: from a small canonical source instance, reduce, solve the target, lift the solution back, and compare against a direct solve of the source (an oracle the rule never invokes).

\textbf{Agentic feature tests.}
Libraries normally stabilize through community use, but a niche library cannot wait for that community to form.
Agentic feature tests approximate user feedback by simulating realistic use scenarios before real users arrive.
The mechanism has three phases.
First, the main agent launches a sub-agent in a \emph{fresh context window} (one that has never seen the implementation) and injects a \emph{user profile}: a structured persona with domain expertise and a concrete use case (Appendix~\ref{app:agent-profile}).
The fresh context is essential; an agent that has just written the code cannot objectively test it.
Second, the sub-agent role-plays the persona end-to-end: it reads only the project's documentation, picks a problem it recognizes from its domain, builds an instance with realistic data, discovers a reduction path through the CLI, solves, and verifies the result against its own expectation.
Third, the sub-agent steps out of character.
The main agent interviews it about what worked, what was confusing, and where the output diverged from expectation, then writes a structured report.
This report surfaces usability gaps, documentation omissions, and semantic errors that unit tests cannot reach.

Agentic feature tests on the \texttt{pred} CLI surfaced $\sim$50 defects in command behavior, help text, and end-to-end pipelines (Appendix~\ref{app:agentic-tests}). Unit and round-trip tests encode the developer's expectations and miss them; agentic feature tests probe the CLI from a fresh user's perspective, exposing documentation drift and overlooked edge cases.
\section{Evidence}\label{sec:evidence}

The pipeline produced a substantive reduction library and an agent-callable interface. 

\subsection{Library Status}

\Cref{fig:overview}(c) summarizes the library at v0.5.0: 190~problem types (220 variants), 265~reduction rules, 129~types (68\%) with a reduction path to ILP, and 78 reachable from 3-SAT.
Each ILP path grants standard solver access via the \texttt{pred} CLI; each 3-SAT path provides a tested NP-hardness argument backed by executable code rather than pen-and-paper proofs.
In 12 weeks of development, most of the library was built after week~8.5, once the full skill-based pipeline was in place (Appendix~\ref{app:phases}).

\subsection{Case Study on End-User Utility}\label{sec:end-user-utility}

The simplest use of \texttt{pred} starts from a high-performance solver and asks what problems it can serve. For an Ising machine that takes QUBO input, \texttt{pred} can list all 138 library problems that reach QUBO within three hops (Appendix~\ref{app:qubo-coverage}). 

The harder direction starts from a problem stated in everyday language and asks an AI agent to use \texttt{pred} to deliver a fast and good solver end-to-end. To check whether \texttt{pred} helps in this direction, we set up a case study where the agent receives a problem in everyday language. In our instance, a manager wants to split a 150-person organization into two product teams to maximize the sum of pair-wise cross-team scores. Mathematically, this is NP-hard Maximum Cut (full setup in Appendix~\ref{app:end-user}).

To answer, the agent has to recognize the underlying mathematical model, pick a reduction path to a solver-friendly endpoint, choose a solver for that endpoint, and write the code that ties them together. Each of these steps depends on prior knowledge that may be wrong, partial, or out of date; \texttt{pred} replaces the reduction-path step with a walk over a verified graph.

Beyond \texttt{pred}, the other obvious way for an agent to reach external information is to search the web. To tell apart \texttt{pred}'s contribution from generic web access, we compare four configurations that toggle the two independently (the four rows of \Cref{tab:end-user-eval}).
Each configuration is run on ten random instances, with one-shot solver wall-time capped at 90~s.
For each run we record three quantities: tokens spent overall, solver wall-time, and the objective value obtained by the solver.

\Cref{tab:end-user-eval} summarizes the outcomes. \emph{With \texttt{pred}, the agent picks the right reduction every time.} Every \texttt{pred}-enabled run lands on the same path (spin-glass / QUBO) and the same solver (simulated annealing on the Ising form), matches the \emph{best-of-four} objective on every instance, and cuts solver wall-time by a factor of $3$--$6$, at the cost of roughly $2{\times}$ more AI tokens (details in Appendix \Cref{tab:end-user-solver-table}).

\emph{Without} \texttt{pred} \emph{, solver choice is unreliable.} The bare configurations scatter over mixed-integer linear programming (MILP), constraint-programming SAT (CP-SAT), simulated annealing, tabu search, and ad-hoc heuristics; the misses always fall on solvers that do not perform well at the 90~s budget. Bare AI misses the best-of-four on $3/10$ instances (worst $19.1\%$ below); bare\,+\,web misses on $2/10$, but its two missed runs fail catastrophically in one-shot execution.

\emph{Web access alone does not close the gap.} It exposes the agent to more candidate solvers but gives no structural guarantee that any of them will work in the budget; on this case study, that broader exposure is net harmful. Appendix~\ref{app:end-user} has more details.

\begin{table}[h]
\caption{Outcomes on the team-split instance class (10~instances per configuration; 90~s solver-wall cap). Agent: Claude Opus~4.7 with extra-high reasoning effort.
Higher objective is better; \emph{Best-match} reports how many of the ten instances each configuration achieves the best-of-four objective.}
\label{tab:end-user-eval}
\centering
\small
\begin{tabular}{@{}lccccc@{}}
\toprule
Configuration & Mean obj.\,$\uparrow$ & Min obj.\,$\uparrow$ & Best-match & Mean AI tokens & Mean solver time (s) \\
\midrule
Bare AI                              & 12{,}306 & 9{,}966 & 7/10  & 37{,}379 & 37.1 \\
Bare AI + web                        & 10{,}014 & 0       & 8/10  & 41{,}504 & 26.1 \\
+ \texttt{pred}                      & 12{,}603 & 12{,}187 & 10/10 & 70{,}523 & 12.9 \\
+ \texttt{pred} + web                & 12{,}603 & 12{,}187 & 10/10 & 80{,}592 &  6.5 \\
\bottomrule
\end{tabular}
\end{table}

\section{Conclusion and Discussion}\label{sec:conclusion}

\subsection{Conclusion}

In about three months, a small team and AI agents built a verified reduction graph of 190 problem types and 265 reduction rules. The resulting library gives those problems verified paths to standard solver targets and executable hardness witnesses.
Building a reduction library at this scale was previously out of reach. What made it tractable is AI coding agents working under the harness we design. The harness loads a project specification into every agent session so contributions compose rather than collide, drives work through role-typed skills, and enforces a layered correctness gate from compile-time checks through unit tests, round-trip tests, agentic feature tests, and final human verification. Speed and rigor reinforce each other in this design: the same harness that accelerates contribution also enforces the checks that prevent drift.

\subsection{Discussion}

\emph{The evolving role of humans.} Harness engineering shifts where human effort is spent, from writing code to designing constraints. The carefully designed harness enforces uniform conventions and quality control across contributors. The most unsettling fact is that none of us is genuinely irreplaceable. We are all replaceable ``plugins'' for responsibilities, creativity, and deep reasoning. Even if a key maintainer stops maintaining the project, a new domain expert can get onboarded quickly with the help of the advisor skills.

\emph{Automation and advisor skills.} Skills can delegate routine work to AI agents. We find them most useful when they guide human contributors. Advisor skills keep humans more actively involved. They support onboarding by lowering the barrier to learning the project's conventions and draw on human expertise for critical decisions. We found this distinction useful when designing the harness. We expect it to become standard vocabulary in harness design.

\emph{Further Directions.}
The reduction graph at scale opens several research directions: one is understanding how computationally hard problems are clustered in problem space; another is identifying high-value missing reduction rule to guide automated reduction-rule discovery.
This reduction graph can be improved in several directions: one is to refine \emph{fine-grained cost prediction} for a given instance rather than rely on asymptotic complexity; another is to extend beyond NP to \emph{other complexity classes} (\#P, PSPACE, etc.).


\bibliographystyle{plainnat}
\bibliography{references}


\appendix

\section{Related Work}\label{sec:related}

\textbf{Reduction libraries and tools.}
Prior work falls along a ladder of executability and scope (\Cref{tab:related-reductions}): paper catalogues and formulation surveys, executable per-rule tools, single-endpoint modeling libraries, and an earlier inter-type prototype. The present library extends this line by connecting \emph{many} source types to \emph{many} targets with composable overhead and inverse maps, under AI-assisted maintenance.

\begin{table*}[h]
\caption{Reduction libraries and tools, ordered by executability and scope.}
\label{tab:related-reductions}
\centering
\small
\setlength{\tabcolsep}{4pt}
\begin{tabular}{@{}p{4.5cm}p{2.6cm}p{6cm}@{}}
\toprule
Work & Form & Scope \\
\midrule
Karp~\cite{karp1972}; Garey and Johnson~\cite{garey1979}; Lucas~\cite{lucas2014} & Paper catalogue & Reduction sketches and Ising formulas; not executable \\
Karp DSL~\cite{Zhang2022Karp} & Executable per-rule & 25 reductions from a textbook; isolated rules, no reusable graph \\
REDNP~\cite{Creus2014REDNP} & Executable per-rule & DSL for student exercises; new problems bootstrap via reduction to SAT \\
PyQUBO~\cite{Zaman2021PyQUBO}; qubovert~\cite{Iosue2022qubovert}; QUBO.jl~\cite{Monteiro2022ToQUBO}; D-Wave Ocean~\cite{DWaveOcean2026} & Executable library & Many problems to single endpoint (QUBO/Ising) \\
ProblemReductions.jl~\cite{gao2025programming,Liu2023GenericTN} & Executable graph & 18 types; inter-type reduction graph; manually maintained \\
\textbf{This work} & Executable graph & 190 types, 265 rules; composable inverse maps; AI-assisted maintenance \\
\bottomrule
\end{tabular}
\end{table*}

Two adjacent traditions warrant mention. Optimization modeling languages such as MiniZinc~\cite{Nethercote2007MiniZinc} let a user write a high-level constraint or optimization model once and dispatch it to many solver back-ends via a shared intermediate representation; the mechanism is modeling-to-solver via flattening rather than problem-to-problem reduction between NP-hard types. Formal verification frameworks, notably the Coq Library of Undecidable Problems~\cite{Forster2020CoqLibrary}, formalize reduction chains whose correctness is machine-checked by proof rather than tested on canonical examples; they target undecidable problems and incur substantially higher development cost than the present library.

\textbf{AI coding agents: capabilities and limitations.}
SWE-agent~\cite{Yang2024SWEagent}, Devin~\cite{Wu2024Devin}, OpenHands~\cite{Wang2024OpenHands}, and Claude Code~\cite{Anthropic2025ClaudeCode} resolve isolated issues with increasing reliability~\cite{Xia2025LiveSWEagent}, but longer-horizon benchmarks reveal a capability cliff: frontier agents resolve up to $\sim$70\% of single-file issues on SWE-bench Verified~\cite{jimenez2024swebench}, yet SWE-bench Pro~\cite{Deng2025SWEBenchPro} reports only $\sim$23\% on multi-file tasks and SWE-EVO~\cite{Thai2025SWEEVO} $\sim$21\%; LongCLI-Bench~\cite{LongCLIBench2026} finds pass rates below 20\% on long-horizon CLI tasks.
Quality concerns are equally sharp: AI co-authored code carries $1.7{\times}$ more major defects~\cite{CodeRabbit2025AICodeQuality}; the METR randomized trial found experienced developers 19\% \emph{slower} with AI tools on their own repositories~\cite{Becker2025METRProductivityRCT}; Cursor's FastRender (3M+ lines generated by parallel agents) scored 1.3/5 for maintainability~\cite{CursorFastRender2026,SIGFastRender2026}.

\textbf{Harness engineering.}
To tame this variance, \emph{harness engineering}~\cite{OpenAI2026HarnessEngineering,Fowler2026HarnessEngineering} shifts the engineer's role from code authorship to environment design.
Persistent project instructions and composable skills~\cite{Claude2026AgentSkills, Obra2025Superpowers} keep agents aligned with repository conventions across sessions; verification pipelines reject incorrect output before it enters the codebase.
Our approach instantiates this pattern for a mathematical domain, exploiting the project's uniform, verifiable structure so that each agent invocation is short, self-contained, and human-reviewable.

\section{CLI Usage}\label{app:cli}

We introduce the \texttt{pred} CLI through one example. The instance is weighted MaxCut on a 4-cycle (\texttt{0-1, 1-2, 2-3, 3-0}) with unit weights. Build the instance:
\begin{quote}\ttfamily\small
pred create MaxCut -{}-graph 0-1,1-2,2-3,3-0 -o instance.json
\end{quote}

\paragraph{Enumerate downstream models.}
\begin{quote}\ttfamily\small
pred from MaxCut/SimpleGraph/i32 -{}-hops 2
\end{quote}
prints the reachable downstream cone (two branches, each three hops from QUBO/f64):
\begin{quote}\ttfamily\small
MaxCut/SimpleGraph/i32\\
\hspace*{1em}|-- MinimumCutIntoBoundedSets/SimpleGraph/i32 -> ILP/bool\\
\hspace*{1em}|-- SpinGlass/SimpleGraph/i32 -> SpinGlass/SimpleGraph/f64
\end{quote}

\paragraph{Compare reduction paths.}
\begin{quote}\ttfamily\small
pred path MaxCut/SimpleGraph/i32 QUBO/f64 -{}-all
\end{quote}
enumerates both paths and reports their per-step overhead. The endpoints differ by orders of magnitude:
\begin{itemize}
\setlength\itemsep{1pt}
\item \emph{Path~1} (via \texttt{SpinGlass}): $\text{num\_vars} = O(|V|)$ at the QUBO endpoint.
\item \emph{Path~2} (via \texttt{MinimumCutIntoBoundedSets} and \texttt{ILP/bool}): $\text{num\_vars} = O(|V|^4)$.
\end{itemize}
Path~2 inflates because the intermediate ILP encoding adds a constraint per cut edge before being quadratised into QUBO; Path~1 keeps the variable count linear. To save Path~1 so that downstream reductions consume the same choice deterministically:
\begin{quote}\ttfamily\small
pred path -{}-cost minimize:num\_vars -o path.json
\end{quote}

\paragraph{Reduce, solve, lift.}
\begin{quote}\ttfamily\small
pred reduce instance.json -{}-via path.json -o bundle.json
\end{quote}
forward-reduces the instance along the saved path; the bundle carries the original instance under \texttt{source}, the reduction trace under \texttt{path}, and the resulting 4-variable QUBO under \texttt{target}. After an external solver returns a target-space configuration (on this instance, alternating colours \texttt{[1,0,1,0]} achieve the QUBO minimum $-8$), the call
\begin{quote}\ttfamily\small
pred extract bundle.json -{}-config 1,0,1,0
\end{quote}
inverse-maps the QUBO solution back through every reduction step and reports the source-space evaluation \texttt{Max(4)} (Appendix~\ref{app:architecture}): every edge of the 4-cycle is cut. \texttt{extract} composes the inverse of every reduction in the path.

\paragraph{Other subcommands.}
\begin{itemize}
\setlength\itemsep{1pt}
\item \texttt{pred show <variant>}: prints a problem's definition, schema, complexity, and neighbouring reductions.
\item \texttt{pred list}: enumerates all registered problem types with their complexities.
\item \texttt{pred evaluate <instance>.json -{}-config <csv>}: scores a candidate configuration against the instance directly, useful for verifying solutions independently of the solver.
\item \texttt{pred solve <instance>.json}: runs an end-to-end solve via the cheapest reduction path with a built-in default solver.
\end{itemize}

\section{Implementation Details}\label{app:architecture}

\begin{figure}[t]
  \centering
  \includegraphics[width=0.5\columnwidth]{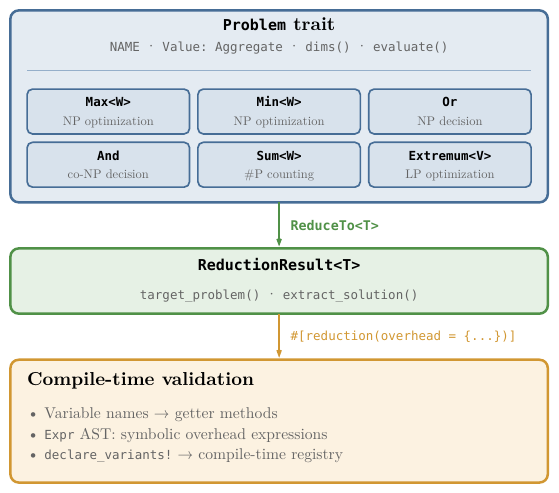}
  \caption{Trait hierarchy and compile-time validation.
    The \texttt{Problem} trait defines a universal evaluation interface; \texttt{ReduceTo<T>} requires both forward and inverse maps; procedural macros validate overhead expressions and variant registrations at compile time.}
  \label{fig:architecture}
\end{figure}

The library's type system reduces the space of possible agent errors by making incorrect code fail to compile.
This appendix describes the key design decisions: the aggregation-based problem representation, the solver priority system, and the variant registry (\Cref{fig:architecture}).

\paragraph{Problems and variants.}
Every problem type implements \texttt{Problem}, which requires: a constant name, an associated \texttt{Value} type, a method returning the configuration space dimensions, an \texttt{evaluate()} method that maps any configuration to a value, and variant metadata.
The \texttt{Value} type is an \emph{aggregation wrapper} that determines how individual evaluations combine across the configuration space.
Six wrappers cover the standard problem classes (\Cref{tab:wrappers}).
Each wrapper implements an \texttt{Aggregate} trait with \texttt{identity()}, \texttt{combine()}, and optional witness-tracking methods, so a brute-force solver reduces to a single fold over the configuration space: \texttt{configs.map(evaluate).fold(identity, combine)}.
Wrappers that support \emph{witness recovery} can track which configuration achieved the optimal value; those that cannot (counting and co-NP) only produce aggregate values.
This distinction propagates to the reduction traits.

\begin{table}[h]
\caption{Aggregation wrappers and their complexity classes.}\label{tab:wrappers}
\centering
\small
\begin{tabular}{@{}llll@{}}
\toprule
Wrapper & Class & Witness & Example \\
\midrule
\texttt{Max<W>} & NP optimization & Yes & Max.\ Indep.\ Set \\
\texttt{Min<W>} & NP optimization & Yes & Min.\ Vertex Cover \\
\texttt{Or}     & NP decision     & Yes & $k$-Satisfiability \\
\texttt{Sum<W>} & \#P counting    & No  & $k$-th Largest $m$-Tuple \\
\texttt{And}    & co-NP decision  & No  & (reserved) \\
\shortstack[l]{\texttt{Extremum}\texttt{<V>}} & NP optimization$^\dagger$ & Yes & Integer Lin.\ Prog. \\
\bottomrule
\end{tabular}
\\[2pt]
{\footnotesize $^\dagger$\texttt{Extremum} differs from \texttt{Max}/\texttt{Min} only in that the optimization sense is chosen at runtime.}
\end{table}
A single problem type admits multiple \emph{variants}: parameterizations by graph type (\texttt{SimpleGraph}, \texttt{PlanarGraph}, \texttt{BipartiteGraph}, \texttt{KingsSubgraph}, \texttt{UnitDiskGraph}, etc.) and weight type (unit, \texttt{i32}, \texttt{f64}).
Each variant may have a better complexity bound: MIS on king's subgraphs runs in $2^{O(\sqrt{|V|})}$~\cite{deBerg2020ETHTight} versus $O(1.1996^{|V|})$ on general graphs~\cite{Xiao2017MIS}.
The \texttt{declare\_variants!} macro registers each concrete instantiation with its best-known complexity, a factory for JSON deserialization, and a solve function.
All variable names in complexity expressions are validated at compile time against the problem's getter methods.
This registry produces the registered variants from 100+ base problem types.

\paragraph{Decision problems.}
A generic \texttt{Decision<P>} wrapper converts any optimization problem \texttt{P} (whose value type is \texttt{Max<W>} or \texttt{Min<W>}) into a decision problem with value type \texttt{Or}: given a bound~$k$, it asks whether a configuration meeting or exceeding~$k$ exists.
The wrapper implements \texttt{Problem} automatically via the \texttt{DecisionProblemMeta} trait, which supplies the decision variant's name.
The reduction from the decision wrapper to its inner optimization type is derived generically, so adding a decision variant requires only a one-line macro invocation (\texttt{register\_decision\_variant!}). 
Three problem types currently have decision variants: Maximum Independent Set, Minimum Vertex Cover, and Minimum Dominating Set.

\paragraph{Reduction rules.}
The type system enforces a split based on witness capability.
\texttt{ReduceTo<T>} requires a \texttt{ReductionResult} that bundles \texttt{target\_problem()} and \texttt{extract\_solution()}, a map from a target \emph{configuration} back to a source configuration.
An agent cannot compile a forward reduction without providing the inverse map; incomplete reductions are compile errors.
\texttt{ReduceToAggregate<T>} requires an \texttt{AggregateReductionResult} with \texttt{extract\_value()} instead, a map from a target \emph{value} back to a source value, with no witness recovered.
Each reduction also carries a compile-time-validated overhead: the \texttt{\#[reduction(overhead = \{...\})]} procedural macro attaches symbolic size expressions, and variable names are checked against getter methods on the source type, so a typo causes a compile error, not a silent bug.

\paragraph{Solvers.}
When the user invokes \texttt{pred solve}, the system selects a solver in priority order.
First, if the problem has a dedicated solver that exploits its structure (currently six problem types), the system uses it.
Second, if a witness-capable reduction path to Integer Linear Programming (ILP) exists, the system reduces along the cheapest path (by Dijkstra) and solves with HiGHS.
Third, the brute-force enumerator is always available: it folds over all configurations using the aggregation interface.
Brute force remains essential for testing, where correctness matters more than speed.

\paragraph{Reduction graph construction.}
Problem variants and reduction rules compose into the directed graph described in \Cref{sec:graph-structure}.
For subtype edges (e.g., MIS on a king's subgraph to MIS on a general graph), a generic \texttt{ReductionAutoCast} implements both reduction traits with an identity solution map.
At runtime, type-erased blanket implementations (\texttt{DynReductionResult}) let the reduction-graph engine chain reductions along Dijkstra-discovered paths without knowing the concrete types at compile time.

\section{QUBO Coverage}\label{app:qubo-coverage}

Specialized hardware for combinatorial optimization is one of the fastest-growing frontiers of AI infrastructure.
Ising machines, quantum annealers~\cite{glover2019}, and physics-inspired neural solvers~\cite{Schuetz2022PhysicsGNN} all promise substantial speedups over conventional CPU solvers, and the field has converged on a single input format: QUBO.
The catch is that the speedup is only available \emph{after} the user has hand-translated their problem into QUBO, a step that has historically required a domain expert per problem and a fresh paper per reduction.
This is exactly where the reduction graph pays off.

Treating QUBO as one registered solver target, \texttt{pred to QUBO -{}-hops 3} reports 138~problem variants that route to it with witness recovery in the current library.
\Cref{tab:qubo-coverage} samples ten and lists the composed overhead $r_{\text{QUBO} \leftarrow A}$, the polynomial mapping source size to QUBO variable count.
All ten share one invocation, \texttt{pred reduce - -{}-to QUBO}, and only the input changes.
By contrast, solver-specific compilers such as PyQUBO~\cite{Zaman2021PyQUBO} and qubovert~\cite{Iosue2022qubovert} accept QUBO input but provide no inter-problem reductions, leaving the translation step to the user.

\begin{table}[htbp]
\caption{Ten problems reaching QUBO within three hops, sampled from 138~reachable variants. A QUBO-capable solver inherits all of them via one command.}
\label{tab:qubo-coverage}
\centering
\small
\begin{tabular}{@{}llll@{}}
\toprule
Problem & Hops & $r_{\text{QUBO} \leftarrow A}$ & Application \\
\midrule
GraphPartitioning           & 1 & $O(V)$              & VLSI partitioning \\
3-Satisfiability            & 1 & $O(C + V)$          & Verification, planning \\
Knapsack                    & 1 & $O(n + b)$          & Resource allocation \\
TravelingSalesman           & 1 & $O(V^2)$            & Logistics, PCB drilling \\
KColoring                   & 1 & $O(V^2)$            & Register allocation \\
Partition                   & 2 & $O(n + b)$          & Load balancing \\
MaximumClique               & 2 & $O(V^3)$            & Bioinformatics \\
BinPacking                  & 2 & $O(n^3)$            & VM/container packing \\
MultiprocessorScheduling    & 2 & $O(P T^2 + P^2 T)$  & Cloud scheduling \\
MaximumIndependentSet       & 3 & $O(V^3)$            & Wireless scheduling \\
\bottomrule
\end{tabular}
\\[2pt]
{\footnotesize $V$: vertices; $C$: clauses; $n$: items; $b$: encoding/slack bits; $P, T$: processors, tasks.}
\end{table}

\section{Skill Excerpts}\label{app:skills}

\Cref{fig:skills} reproduces the two skills referenced in \Cref{sec:pipeline-overview}: \texttt{propose} (advisor skill that gates issue creation by domain experts) and \texttt{add-rule} (automation skill that drives implementation from a validated issue).

\begin{figure*}[htbp]
  \centering
  \begin{minipage}[t]{0.525\textwidth}
  \begin{tcolorbox}[
    colback=gray!5, colframe=gray!80,
    title={\textbf{\texttt{propose}} --- advisor skill (excerpts)},
    fonttitle=\footnotesize, boxrule=0.4pt, arc=2pt, left=4pt, right=4pt, top=2pt, bottom=2pt,
    equal height group=skillbox,
  ]
  \footnotesize
  \begin{lstlisting}[basicstyle=\ttfamily\scriptsize, numbers=none, frame=none, aboveskip=0pt, belowskip=0pt, xleftmargin=4pt]
---
name: propose
description:
  Guides domain experts through brainstorm-
  ing and files a GitHub issue
---
  \end{lstlisting}
  \vspace{-4pt}
  \rule{\linewidth}{0.3pt}\\[3pt]
  \textbf{No programming knowledge required.}
  This skill works entirely in mathematical / domain language.\\[3pt]
  \texttt{<HARD-GATE>}\\
  Do NOT write any code, create any files, or invoke implementation skills.
  The ONLY output of this skill is GitHub issues filed via \texttt{gh issue create}.\\
  \texttt{</HARD-GATE>}\\[3pt]
  \rule{\linewidth}{0.3pt}\\[3pt]
  \textbf{Run topology analysis first} to identify the most impactful missing reductions [\ldots]\\
  \begin{lstlisting}[basicstyle=\ttfamily\scriptsize, numbers=none, frame=none, aboveskip=-5pt, belowskip=2pt, xleftmargin=4pt]
cargo run --example \
    detect_isolated_problems
cargo run --example \
    detect_unreachable_from_3sat
  \end{lstlisting}
  \vspace{-2pt}
  \rule{\linewidth}{0.3pt}\\[3pt]
  \textbf{Pre-validation} --- Run \texttt{check-issue} on draft before filing:\\[2pt]
  \begin{tabular}{@{}r@{~}l@{}}
    1. & \textbf{Usefulness} --- does a cheaper reduction path already exist? \\
    2. & \textbf{Effort} --- effort to implement? \\
    3. & \textbf{Correctness} --- do cited references support the claims? \\
    4. & \textbf{Writing quality} --- all symbols defined, examples worked? \\
  \end{tabular}\\[2pt]
  Only issues passing all four checks are filed.
  \end{tcolorbox}
  \label{fig:skill-propose}
  \end{minipage}\hspace{4pt}%
  \begin{minipage}[t]{0.445\textwidth}
  \begin{tcolorbox}[
    colback=gray!5, colframe=gray!80,
    title={\textbf{\texttt{add-rule}} --- automation skill (excerpts)},
    fonttitle=\footnotesize, boxrule=0.4pt, arc=2pt, left=4pt, right=4pt, top=2pt, bottom=2pt,
    equal height group=skillbox,
  ]
  \footnotesize
  \begin{lstlisting}[basicstyle=\ttfamily\scriptsize, numbers=none, frame=none, aboveskip=0pt, belowskip=0pt, xleftmargin=4pt]
---
name: add-rule
description: Adding a new reduction 
  rule to the codebase, either from 
  an issue or interactively
---
  \end{lstlisting}
  \vspace{-4pt}
  \rule{\linewidth}{0.3pt}\\[3pt]
  \textbf{Required Information Checklist}
  \begingroup
  \setlength{\leftmargini}{1.4em}
  \setlength{\itemsep}{1pt}
  \setlength{\parskip}{0pt}
  \begin{enumerate}
    \item \textbf{Reduction algorithm} --- how to transform source instance to target, with BibTeX reference.
    \item \textbf{Size overhead} --- how target size relates to source size, used for estimating the overhead of the reduction.
    \item \textbf{Concrete example} --- a small worked-out instance, used for testing and documentation.
    \item $\ldots$
  \end{enumerate}
  \endgroup
  Do NOT proceed until the checklist is complete.\\[3pt]
  \rule{\linewidth}{0.3pt}\\[3pt]
  \textbf{Reference Implementations} --- Read these first:\\
  {\ttfamily src/rules/mvc\_mis.rs} [\ldots{}]\\[3pt]
  %
  \rule{\linewidth}{0.3pt}\\[3pt]
  \textbf{Document in paper.}
  Write a \texttt{reduction-rule} entry in the Typst paper with theorem body, proof (construction, correctness, solution extraction), and a worked example with concrete numbers.
  Run \texttt{make paper} --- must compile without errors.
  \end{tcolorbox}
  \label{fig:skill-addrule}
  \end{minipage}
  \caption{Two skills that bookend the contribution pipeline.
    \textbf{\texttt{propose}} (left) gates issue creation by domain experts: the hard gate enforces no-code contribution, topology analysis guides experts toward strategic reduction-graph gaps, and pre-validation checks four quality dimensions before the issue enters the pipeline.
    \textbf{\texttt{add-rule}} (right) drives implementation from a validated issue: the checklist encodes the mathematical knowledge every reduction requires, reference implementations prevent convention drift, and the paper entry ensures visual verifiability by human reviewers.}
  \label{fig:skills}
\end{figure*}

\section{Agent Profile Example}\label{app:agent-profile}

\Cref{fig:agent-profile} shows a representative agent profile used by the \emph{agentic feature tests} described in \Cref{sec:verification}. The persona's domain knowledge lets the sub-agent judge whether reduction outputs are plausible without access to ground-truth labels.

\begin{figure}[h]
  \centering
  \begin{tcolorbox}[
    colback=gray!5, colframe=gray!80,
    title={\textbf{Agent Profile:} Dr.\ Sarah Chen --- CLI Tool (\texttt{pred})},
    fonttitle=\small, boxrule=0.4pt, arc=2pt, left=4pt, right=4pt, top=2pt, bottom=2pt
  ]
  \small
  \textbf{Use case.}
  An algorithm engineer reads the project documentation to learn what problems and reductions exist, then uses the \texttt{pred} CLI to model and solve their own optimization problems.
  Each session: pick a problem from the paper, create an instance with real-world-style data, find a reduction path to a solver, solve it, and verify the result.\\[4pt]
  \textbf{Persona.}
  Senior algorithm engineer at a logistics company with ten years of experience in vehicle routing, scheduling, and layout optimization.
  Knows these problems map to graph coloring, independent set, set cover, and similar NP-hard formulations because she has hand-coded reductions before.
  Found this project's paper and wants to try the CLI instead of writing reductions herself.
  Has never used \texttt{pred} before.\\[4pt]
  \textbf{Behavior.}
  Reads the paper and docs first, then picks a problem she recognizes.
  Figures out the CLI by reading its help output.
  After each step, compares the output against her expectation.
  If something does not match, investigates before moving on.
  \end{tcolorbox}
  \caption{A representative agent profile used in agentic feature testing.
    The persona's domain knowledge (logistics optimization) lets the agent judge whether reduction results are plausible without access to ground-truth labels.}
  \label{fig:agent-profile}
\end{figure}

\section{Agentic Feature Test Findings}\label{app:agentic-tests}

The ${\sim}\,50$ \texttt{pred} defects of \Cref{sec:verification} fall into two categories. Every defect is faithfully recorded in the project's development history (GitHub issues and pull requests), which will be made public alongside the paper.

\paragraph{Help text drifts from actual behaviour.}
A user only knows what the program tells them, so any disagreement between the documentation and the running code propagates into every later decision the user makes. As one example, in \texttt{pred} v0.3.0 typing \texttt{pred} with no subcommand printed the top-level menu and then appended roughly seventy lines of unrelated \texttt{pred solve} help, which may lead the user to believe a solve invocation is required by default. A simulated first-time user found this on the very first command and reported the confusion. An AI agent writing unit tests thinks about the code it has just built, not about how a user would first meet the tool.

\paragraph{Counterexamples and edge cases the test author missed.}
Verifying every reduction rule without formal proof or automated theorem proving has no clean solution. The project mitigates this in two layers. The first layer frontloads correctness: domain experts file high-quality issues drawn from their experience and from canonical references (like Garey and Johnson~\cite{garey1979}), and each problem type ships with a non-trivial canonical example chosen to trigger the structural features the reductions must respect. Current AI cannot reliably construct such examples (\Cref{sec:human-responsibilities}); it defaults to toy cases that satisfy the type contract but miss the structure that makes the reduction non-trivial. That structure is difficult to articulate in language, and AI can be nudged toward less trivial cases only through explicit proxies such as variable count or edge density. The second layer is agentic feature tests, which run the real CLI pipeline on those canonical examples from the outside and catch contract violations that the existing tests miss. As one example, an instance of an NP-hard problem typically has many distinct valid solutions, and a reduction's inverse map must lift any of them back to a valid source solution; unit tests check this property using only the one solution a brute-force solver happens to return, and an agentic-test sweep caught four reductions whose inverse maps worked there but silently produced \emph{invalid} source configurations when the CLI's default ILP solver returned a different valid solution. As another, the \texttt{-{}-cost-bound} flag of \texttt{pred create ProductionPlanning} was declared as a signed integer but cast to an unsigned integer without a non-negativity check, so a simulated user typing a negative bound silently received a problem with an effectively unbounded cost, which the solver declared trivially feasible. Agentic feature tests do not remove this intrinsic difficulty, but by probing for counterexamples from a user's perspective they progressively harden the library against the inputs that real use produces.

Both categories share an origin: agentic feature tests evaluate the program from the outside, using only the documentation and realistic data, and find what an unguarded user would find.

\section{Development Phases}\label{app:phases}

In Phase~1 (weeks~0--7), the team and AI agents ported reduction rules from existing packages (ProblemReductions.jl~\cite{gao2025programming}, UnitDiskMapping.jl~\cite{Nguyen2023UnitDiskMapping,pan2025encodingcomputationallyhardproblems}, and qubogen~\cite{Tamura2020qubogen}) and from Garey and Johnson's catalogue~\cite{garey1979}.
Growth was steady but slow: roughly 20~problem types and 55~rules by week~7.
Through this process, the skill-based automation pipeline was established and iteratively refined.
In Phase~2 (weeks~7--8.5), basic skills were introduced, but the pipeline was not yet end-to-end automated.
In Phase~3 (weeks~8.5--13), the full pipeline became operational.
Both curves accelerate sharply: problem types grew from 23 to 187, and reduction rules from 52 to 239, in under five weeks.
External domain experts began contributing through the \texttt{propose} skill, filed structured issues in mathematical language while agents handled implementation.

\section{End-User Utility Experiment Details}\label{app:end-user}

This appendix expands on the controlled experiment summarised in \Cref{sec:end-user-utility}.

\paragraph{Instance generator.}
Each of the ten seeds (the primes 7, 11, 13, 17, 19, 23, 29, 31, 37, 41) deterministically produces one team-split scenario: 150 engineers with a dense signed-weight survey over the 150-choose-2 pair graph. Each pair is independently sampled with probability $0.5$ and, if sampled, receives an integer weight drawn uniformly from $[-5, 12]$ (zeros skipped). Positive weights mean the pair benefits from being on different teams; negative weights mean they belong on the same team. The resulting graph has roughly 5{,}300 edges (density ${\approx}\,0.48$) with mixed signs. The corresponding mathematical model is signed-weight Maximum Cut on a 150-vertex dense graph.

\paragraph{Four configurations.}
The agent runs in one of four configurations differing on two binary axes ($2{\times}2$ factorial over \emph{web search} and \texttt{pred} \emph{access}):
\begin{itemize}
\item \textbf{Bare AI (A):} no \texttt{pred}, no web. The agent works from training-time priors only.
\item \textbf{Bare AI + web (B):} no \texttt{pred}, but \texttt{WebSearch}/\texttt{WebFetch} allowed.
\item \textbf{+ \texttt{pred} (C):} \texttt{pred} CLI available, no web.
\item \textbf{+ \texttt{pred} + web (D):} both \texttt{pred} and web available.
\end{itemize}
Each cell runs ten seeds under the same environment settings (\emph{Solver environment} below); A and B receive only the user-facing prompt, while C and D additionally receive a Markdown skill (\texttt{solve-instance-offline} for C, \texttt{solve-instance} for D).

\paragraph{Scope of \texttt{pred}'s contribution (C and D).}
In C and D the only thing we add on top of A and B is a Markdown skill explaining how to use the \texttt{pred} CLI to walk the reduction graph; the skill says nothing about which solver to run. It guides the agent through two steps. The first is reduction-path selection: match the user's problem description to a known model, query \texttt{pred} for downstream endpoints ranked by overhead and worst-case complexity, save a chosen path, forward-reduce along it, and inverse-map the solver's answer back to the original variables (Appendix~\ref{app:cli} shows an example). The second is solver selection: given the endpoint \texttt{pred} returned, pick an off-the-shelf solver that handles instances of that size and structure within the time budget. The skill lists no candidate solvers and expresses no preference; C relies on the agent's training-time knowledge for this step, and D additionally uses \texttt{WebSearch}/\texttt{WebFetch}. This split keeps the comparison fair: \texttt{pred} only helps with picking the reduction endpoint, and the choice of solver on top of that endpoint is left entirely to the agent.

\paragraph{Solver environment.}
Each run uses a fresh working directory and a fresh Python virtual-environment created via \texttt{uv}; the agent installs whatever solver package it picks. All four configurations follow the same single-pass protocol: the agent commits to one algorithm and one hyperparameter configuration, invokes the solver exactly once as \texttt{timeout 90 python solve.py}, and reports whatever target-space configuration the call returns (no retries, no method chaining, no polish stage). Each run is handled by Claude Opus~4.7 at extra-high reasoning effort (\texttt{claude-opus-4-7-xhigh}) in its own workdir.

\paragraph{User-facing prompt.}
All four configurations receive the same engineer-voice prompt rendered from a shared template (\Cref{fig:end-user-prompt}). The template carries no model or solver hint. The placeholders \texttt{\{NUM\_PEOPLE\}}, \texttt{\{LAST\_ID\}}, and \texttt{\{NUM\_PAIRS\}} are interpolated per seed.

\begin{figure}[h]
\centering
\begin{tcolorbox}[colback=gray!5, colframe=gray!80, title={\textbf{User-facing prompt template (rendered identically for all four configurations)}}, fonttitle=\footnotesize, boxrule=0.4pt, arc=2pt, left=4pt, right=4pt, top=2pt, bottom=2pt]
\begin{lstlisting}[basicstyle=\ttfamily\scriptsize, numbers=none, frame=none, aboveskip=0pt, belowskip=0pt, breaklines=true, xleftmargin=4pt]
Hi - I run people-ops for a {NUM_PEOPLE}-person engineering org and we're
about to do our annual team re-org. The CTO wants to split the whole org
into exactly two product teams (call them Team A and Team B), and she
wants the split chosen to maximize total cross-team collaboration value.

Here's the model:
We have {NUM_PEOPLE} engineers (id 0..{LAST_ID}). HR has run a yearly
survey asking everyone about who they've worked with productively, who
they've clashed with, and so on. From that I built a per-pair score:

  - score > 0 means putting these two on DIFFERENT teams is GOOD.
  - score < 0 means they're better off on the SAME team.
  - pairs not in the CSV have score = 0.

The objective we're maximizing is:

    sum over all listed pairs of   score_pair,
    ONLY IF the two people end up on different teams.

There is no team-size constraint - Team A can be any size from 0 to

{NUM_PEOPLE}, and Team B gets the rest. We just want the split that
maximizes the cross-team value sum.

I've attached the per-pair scores as `pairs.csv` in your workdir
({NUM_PAIRS} rows total - the survey is dense). Each row is
`a,b,score` (3 columns, no header), with engineer ids in 0..{LAST_ID}.

Please write a script that picks the split and prints the resulting total
cross-team value. Save the answer as `answer.json` with structure:

    { "team_a": [3, 7, 19, 42, ...] }

I'd like the algorithm to be both fast and good. Thanks!
\end{lstlisting}
\end{tcolorbox}
\caption{Shared user-facing prompt rendered to all four configurations.}
\label{fig:end-user-prompt}
\end{figure}

\paragraph{Per-instance solver picks.}
\Cref{tab:end-user-solver-table} lists the solver each agent selected on every instance.
Without \texttt{pred}, the bare baselines drift across MILP (HiGHS, PuLP+CBC), OR-Tools CP-SAT~\cite{Perron2024ORTools}, simulated annealing (\texttt{dwave-neal} / \texttt{dwave-samplers}~\cite{DWaveOcean2026}), \texttt{dwave-tabu}~\cite{DWaveOcean2026}, and Goemans--Williamson~\cite{GoemansWilliamson1995} (PICOS+CVXOPT). The two failure modes show why \texttt{pred} is non-trivial. A's misses are MILP and CP-SAT runs whose LP relaxation is too loose for branch-and-bound to close in 90~s. B's misses are sharper: prompted by web search, the agent picks fancier methods (tabu on seed 13, SDP on seed 29) and trips on integration details that documentation snippets do not surface. The tabu run misreads a solver-parameter convention and ends up asking for a compute budget orders of magnitude beyond the 90~s wall, so the call is killed before any answer is returned. The SDP run pairs Goemans--Williamson with a free open-source backend (CVXOPT) that cannot finish on a problem of this size in 90~s, where a commercial backend (MOSEK, SDPA) would have converged. In both cases the chosen method is sound on this problem; what fails is matching it to a working API call and a backend that scales. \texttt{pred} addresses precisely this gap. The fundamental difference is in the search space the agent navigates: without \texttt{pred}, the agent picks from any solver web search surfaces, including fancier methods whose APIs and backends are easy to misuse; with \texttt{pred}, the reduction graph routes the problem to a target form (here, the Ising / QUBO family) whose standard solvers are mature and well-integrated. The agent still picks its own solver on top of that endpoint, but the choice is now made in a region where the integration is well-trodden by construction. A natural extension is to annotate each endpoint in the reduction graph with richer solver-level information (recommended packages, working API patterns, known backend choices for given problem sizes), narrowing the agent's remaining choice further still.

With \texttt{pred} (C and D), every run reduces the problem to QUBO via the SpinGlass route and solves it with simulated annealing on the Ising form, matching the best-of-four on all ten instances. \texttt{pred} makes this route the default rather than one option among many. Its overhead-aware ranking ($O(|V|)$ for SpinGlass vs.\ $O(|V|^4)$ for the ILP route) surfaces the SpinGlass endpoint, where signed-weight MaxCut maps directly to an Ising Hamiltonian (each edge weight becomes a coupling) without auxiliary variables. On that form simulated annealing's local spin flips converge to low-energy configurations well inside the 90~s budget for a 150-variable problem. 

\begin{table}[h]
\caption{Solver picked on each instance by each configuration.
SA~=~simulated annealing on the Ising / QUBO reduction (\texttt{dwave-neal} or \texttt{dwave-samplers}); MILP~=~mixed-integer linear program (HiGHS or PuLP+CBC); CP-SAT~=~OR-tools constraint programming; Tabu~=~D-Wave \texttt{TabuSampler}; SDP~=~Goemans--Williamson semidefinite relaxation (PICOS+CVXOPT).
Bold cells are the largest gaps to the best-of-four.}
\label{tab:end-user-solver-table}
\centering
\small
\begin{tabular}{@{}c@{\hspace{2pt}}llll@{}}
\toprule
Seed & Bare AI         & Bare AI + web   & + \texttt{pred}   & + \texttt{pred} + web \\
\midrule
 7 & SA                & SA                & SA & SA \\
11 & SA                & SA                & SA & SA \\
13 & \textbf{CP-SAT}   & \textbf{Tabu}     & SA & SA \\
17 & \textbf{MILP}     & SA                & SA & SA \\
19 & SA                & SA                & SA & SA \\
23 & SA                & SA                & SA & SA \\
29 & SA                & \textbf{SDP}      & SA & SA \\
31 & SA                & SA                & SA & SA \\
37 & \textbf{MILP}     & SA                & SA & SA \\
41 & SA                & SA                & SA & SA \\
\midrule
\textbf{solver families} & 3 & 3 & 1 & 1 \\
\bottomrule
\end{tabular}
\end{table}

\paragraph{Raw per-run data.}
\Cref{tab:end-user-raw} lists the objective and corresponding solver wall-time for each (configuration, instance) pair. Bold objectives mark misses below the best-of-four.

\begin{table*}[h]
\caption{Raw per-instance objective and solver wall-time for each configuration. Bold objectives fall below the best-of-four (right column). The 90~s OS-enforced kill cutoff bounds wall-times; entries near 80--90~s indicate the solver was running until killed.}
\label{tab:end-user-raw}
\centering
\footnotesize
\setlength{\tabcolsep}{4pt}
\begin{tabular}{@{}c cc cc cc cc c@{}}
\toprule
& \multicolumn{2}{c}{Bare AI} & \multicolumn{2}{c}{Bare AI + web} & \multicolumn{2}{c}{+ \texttt{pred}} & \multicolumn{2}{c}{+ \texttt{pred} + web} & \\
\cmidrule(lr){2-3} \cmidrule(lr){4-5} \cmidrule(lr){6-7} \cmidrule(lr){8-9}
Seed & Obj. & Time (s) & Obj. & Time (s) & Obj. & Time (s) & Obj. & Time (s) & Best of 4 \\
\midrule
 7 & 12{,}962          & 13.0 & 12{,}962          & 25.3 & 12{,}962 &  1.3 & 12{,}962 &  1.2 & 12{,}962 \\
11 & 12{,}641          &  6.5 & 12{,}641          &  6.4 & 12{,}641 & 15.3 & 12{,}641 & 11.9 & 12{,}641 \\
13 & \textbf{12{,}552} & 75.0 & \textbf{0}        & 90.0 & 12{,}944 &  6.1 & 12{,}944 & 11.7 & 12{,}944 \\
17 & \textbf{9{,}966}  & 83.0 & 12{,}326          &  3.2 & 12{,}326 &  2.9 & 12{,}326 &  1.2 & 12{,}326 \\
19 & 12{,}915          & 12.5 & 12{,}915          &  3.0 & 12{,}915 & 14.7 & 12{,}915 &  5.9 & 12{,}915 \\
23 & 12{,}240          & 12.1 & 12{,}240          &  6.3 & 12{,}240 & 59.7 & 12{,}240 &  1.2 & 12{,}240 \\
29 & 12{,}944          & 25.0 & \textbf{0}        & 90.0 & 12{,}944 & 12.6 & 12{,}944 &  5.8 & 12{,}944 \\
31 & 12{,}401          & 32.8 & 12{,}401          &  1.3 & 12{,}401 &  3.0 & 12{,}401 & 11.6 & 12{,}401 \\
37 & \textbf{12{,}250} & 85.1 & 12{,}471          & 32.5 & 12{,}471 & 12.5 & 12{,}471 & 14.4 & 12{,}471 \\
41 & 12{,}187          & 26.2 & 12{,}187          &  3.4 & 12{,}187 &  1.2 & 12{,}187 &  0.6 & 12{,}187 \\
\midrule
\textbf{mean} & 12{,}306 & 37.1 & 10{,}014 & 26.1 & \textbf{12{,}603} & \textbf{12.9} & \textbf{12{,}603} & \textbf{6.5} & 12{,}603 \\
\bottomrule
\end{tabular}
\end{table*}

\paragraph{Role of web access.}
Comparing A and B isolates the effect of web access on the bare configurations. Web has real but two-sided effects: it can steer the agent away from exact methods like MILP that do not scale to this problem size (B's match rate rises to $8/10$ from A's $7/10$), but the broader candidate pool also exposes the agent to fancier methods whose integrations are subtle, and two of B's runs return $0$ (dragging its mean to $10{,}014$, below A's $12{,}306$). Web changes which solver the agent picks but provides no structural guarantee about budget feasibility; the net effect on this case study is unreliable. The agent still needs domain-specific tools and skills (here \texttt{pred}) to reliably pick an effective solver.

On C and D every run lands on simulated annealing (SA), so the only remaining decision is the SA budget. \Cref{tab:end-user-sa-hyperparams} shows that the two configurations pick similar budgets on most seeds, but D's largest budgets stay well below C's. The reason shows up in D's logs: D queries WebSearch for typical \texttt{dwave-neal} budgets and anchors its choices near the documented values, while C, lacking that reference, occasionally inflates the budget as a safety margin.

\begin{table}[!ht]
\caption{Simulated-annealing budget chosen by each \texttt{pred}-enabled run and the resulting solver wall-time. Each cell shows \texttt{num\_reads} $\times$ \texttt{num\_sweeps}.}
\label{tab:end-user-sa-hyperparams}
\centering
\small
\begin{tabular}{@{}ccccc@{}}
\toprule
       & \multicolumn{2}{c}{+ \texttt{pred}} & \multicolumn{2}{c}{+ \texttt{pred} + web} \\
\cmidrule(lr){2-3} \cmidrule(lr){4-5}
Seed   & reads $\times$ sweeps      & time (s) & reads $\times$ sweeps      & time (s) \\
\midrule
 7 & $200 \times 2{,}000$           &  1.3 & $200 \times 2{,}000$           &  1.2 \\
11 & $1{,}000 \times 5{,}000$       & 15.3 & $400 \times 10{,}000$          & 11.9 \\
13 & $200 \times 10{,}000$          &  6.1 & $400 \times 10{,}000$          & 11.7 \\
17 & $1{,}000 \times 1{,}000$       &  2.9 & $200 \times 2{,}000$           &  1.2 \\
19 & $1{,}000 \times 5{,}000$       & 14.7 & $200 \times 10{,}000$          &  5.9 \\
23 & $2{,}000 \times 10{,}000$      & 59.7 & $200 \times 2{,}000$           &  1.2 \\
29 & $200 \times 20{,}000$          & 12.6 & $200 \times 10{,}000$          &  5.8 \\
31 & $200 \times 5{,}000$           &  3.0 & $200 \times 20{,}000$          & 11.6 \\
37 & $200 \times 20{,}000$          & 12.5 & $500 \times 10{,}000$          & 14.4 \\
41 & $200 \times 2{,}000$           &  1.2 & $200 \times 1{,}000$           &  0.6 \\
\bottomrule
\end{tabular}
\end{table}

Taken together, A vs B and C vs D give a consistent picture of what web access does and does not do in one-shot solving. Web has real effects in both directions. Without \texttt{pred}'s structural guardrail, those effects are unreliable: web changes which solver the agent picks, sometimes for the better and sometimes catastrophically for the worse. With \texttt{pred} fixing the solver class, web takes on a narrower role: it can serve as a documentation lookup that anchors parameter choices to known-good ranges. This split reflects a more general fact: there are few universal priors for solver parameters; choices are usually case-by-case, and even experienced practitioners reach good settings through grid search over plausible values rather than a single recipe. \texttt{pred} contributes on the solver-selection axis, where structure provides a reliable criterion; tuning a chosen solver's parameters is naturally layered on top as an iterative grid-search refinement.


\clearpage

\end{document}